\documentclass[10pt,twocolumn,letterpaper]{article}
\pdfoutput=1

\usepackage{cvpr}              
\usepackage[dvipsnames]{xcolor}

\usepackage[accsupp]{axessibility}
\definecolor{cvprblue}{rgb}{0.21,0.49,0.74}
\definecolor{dark-green}{rgb}{0,0.5,0}
\definecolor{chromeyellow}{rgb}{1.0, 0.65, 0.0}

\usepackage{hyperref}
\hypersetup{pagebackref,breaklinks,colorlinks,citecolor=cvprblue}
\usepackage{multirow}
\usepackage{arydshln}
\usepackage{utfsym}
\usepackage{colortbl}

\title{VRP-SAM: SAM with Visual Reference Prompt}

\author{
Yanpeng Sun$^{1,2*}$, \qquad Jiahui Chen$^{2, 3}$\thanks{equal contribution},\qquad Shan Zhang$^4$,\qquad Xinyu Zhang$^2$,\qquad Xiaofan Li$^2$\\ 
Qiang Chen$^2$,\qquad Gang Zhang$^2$,\qquad Errui Ding$^2$,\qquad Jingdong Wang$^2$,\qquad  Zechao Li$^1$\thanks{Corresponding author.}\\[3mm]
$^1$Nanjing University of Science and Technology,\\
$^2$Baidu VIS,\qquad $^3$Beihang University,\qquad $^4$Australian National University\\
{\tt\small \{yanpeng\_sun, zechao.li\}@njust.edu.cn}
}
\begin{document}
\maketitle
\begin{abstract} 
In this paper, we propose a novel Visual Reference Prompt (VRP) encoder that empowers the Segment Anything Model (SAM) to utilize annotated reference images as prompts for segmentation, creating the VRP-SAM model. In essence, VRP-SAM can utilize annotated reference images to comprehend specific objects and perform segmentation of specific objects in target image. It is note that the VRP encoder can support a variety of annotation formats for reference images, including \textbf{point}, \textbf{box}, \textbf{scribble}, and \textbf{mask}. VRP-SAM achieves a breakthrough within the SAM framework by extending its versatility and applicability while preserving SAM's inherent strengths, thus enhancing user-friendliness. To enhance the generalization ability of VRP-SAM, the VRP encoder adopts a meta-learning strategy. To validate the effectiveness of VRP-SAM, we conducted extensive empirical studies on the Pascal and COCO datasets. Remarkably, VRP-SAM achieved state-of-the-art performance in visual reference segmentation with minimal learnable parameters. Furthermore, VRP-SAM demonstrates strong generalization capabilities, allowing it to perform segmentation of unseen objects and enabling cross-domain segmentation. The source code and models will be available at \url{https://github.com/syp2ysy/VRP-SAM}

\end{abstract}    
\pdfoutput=1
\section{Introduction}

In recent, the Segment Anything Model (SAM)~\cite{sam} has emerged as a foundational visual model for image segmentation. Trained on an extensive datasets comprising billions of labels, SAM has demonstrated remarkable versatility in the realm of universal segmentation. What sets SAM apart is its human-interactive design, allowing segmentation based on user-provided prompts, be they in the form of points, bounding boxes, or coarse masks. This distinctive feature positions SAM as a robust tool that can be adaptable to various tasks and requirements~\cite{hqsam,seem}.

\begin{figure}[t]
\centering
\includegraphics[width=0.47\textwidth]{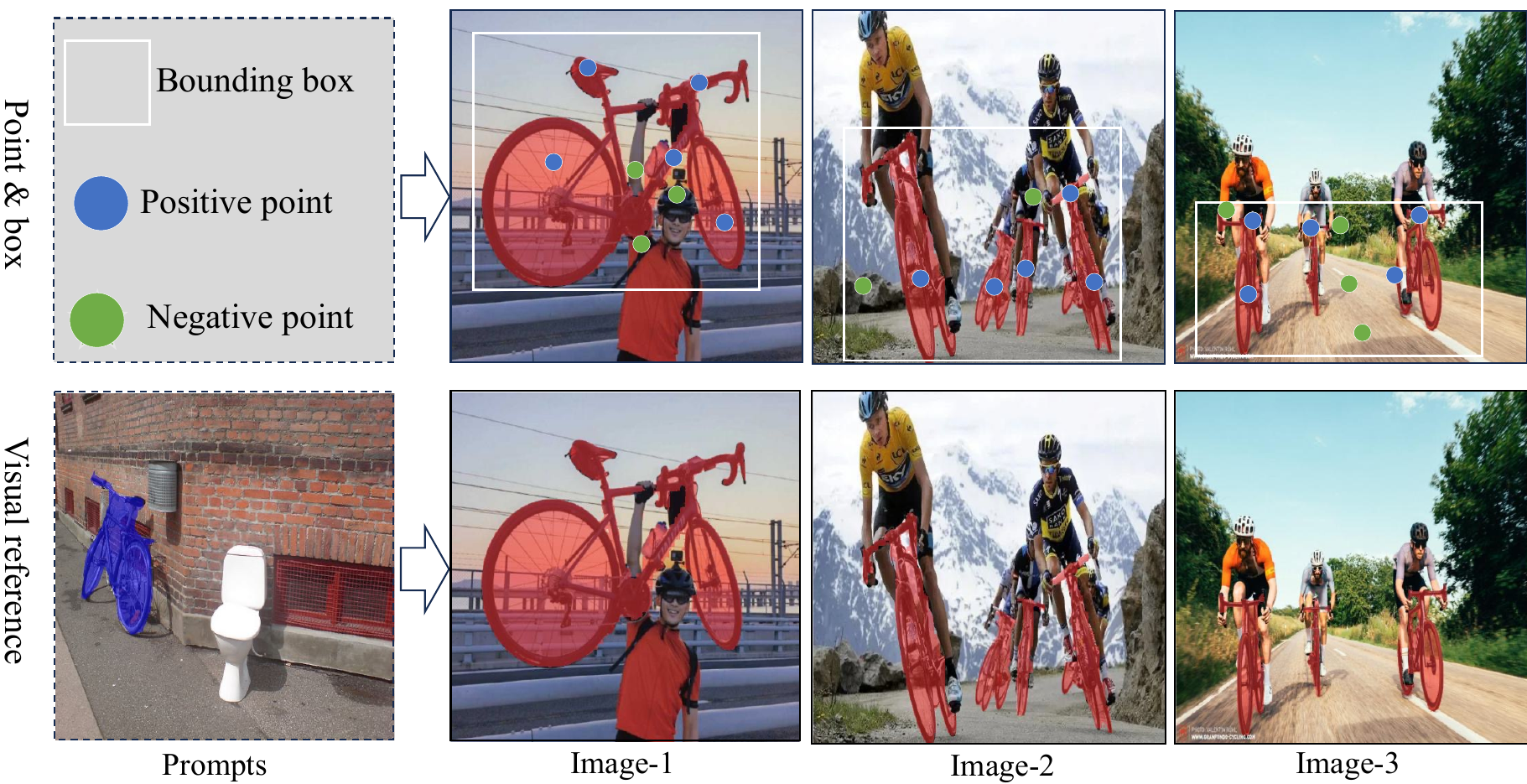} 
\caption{Comparison of SAM's built-in Point and Box prompt modes with Visual Reference Prompt when handling numerous images. The prompts are all provided by users.}
\label{fig:vs_prompts}
\vspace{-1.5em}
\end{figure}

\begin{figure}[t]
\centering
\includegraphics[width=0.47\textwidth]{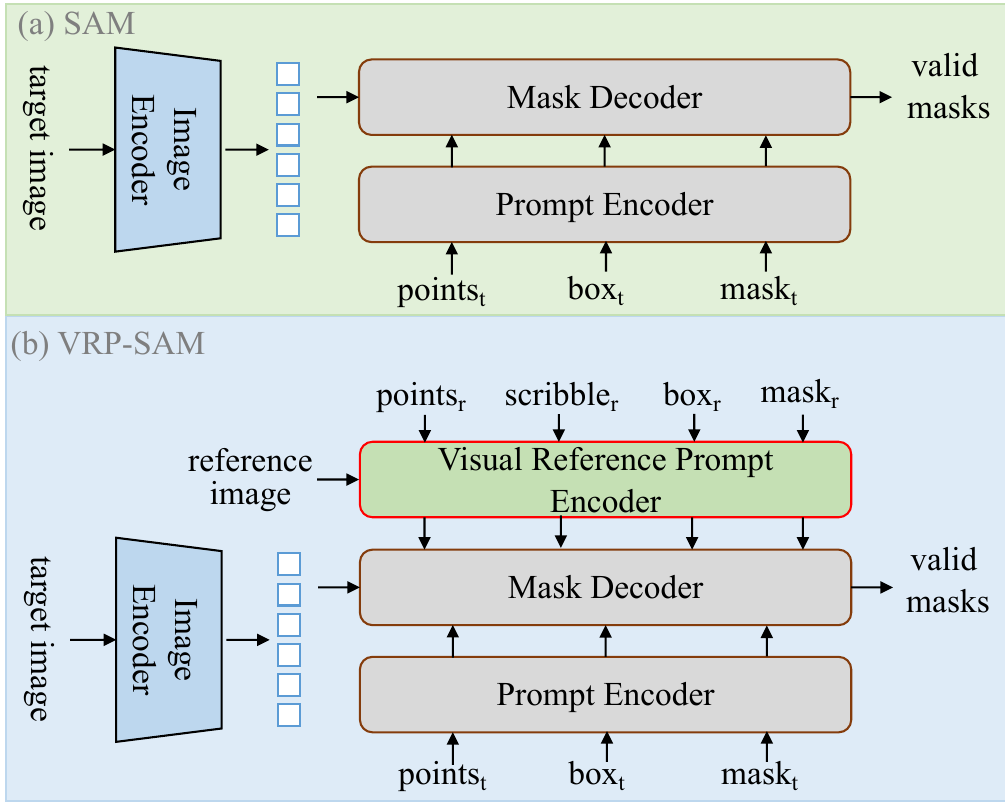} 
\caption{A Comparison between SAM and VRP-SAM. VRP-SAM introduces a visual reference prompt encoder, accepting annotated reference images with \textbf{point}, \textbf{scribble}, \textbf{box}, and \textbf{mask} formats, offering a distinct enhancement over SAM.}
\label{fig:vs_sam}
\vspace{-1.5em}
\end{figure}

However, the existing prompt formats of SAM present significant challenges in practical applications, especially when dealing with complex scenes and numerous images. As shown in Figure~\ref{fig:vs_sam}(a), SAM relies on user-provided prompts (points, boxes, coarse mask) to segment objects in the target image, demanding users to possess a comprehensive understanding of the target objects. In real-world applications, especially in complex scenarios, the level of user familiarity with the target objects can significantly impact the effectiveness of providing specific prompts. Furthermore, variations in the position, size, and quantity of target objects across different images require custom prompts for each image. As illustrated in Figure~\ref{fig:vs_prompts} with the aim of segmenting \textit{'bicycles'}, users need to customize different prompts for each image, significantly impacting efficiency of SAM. Therefore, we propose integrating visual reference prompts to overcome these limitations and enhance adaptability of SAM.

Visual reference prompts is annotated reference images that delineate the objects users expect to segment. As shown in Figure~\ref{fig:vs_prompts}, by simply providing a visual reference prompt with \emph{bicycle}, we can segment \emph{bicycle} in different images without requiring users to provide specific prompts for each image. It significantly enhances the efficiency of SAM while reducing reliance on user familiarity with objects. To achieve this goal, some methods~\cite{matcher,persam} incorporate semantic correlation models~\cite{dinov2,he2016deep} to establish reference-target correlation~\footnote{SAM, being a category-agnostic model, encounters challenges in adequately capturing reference-target correlations.}, obtaining pseudo-masks for the target objects. Following this, a sampling strategy is devised to extract a set of points and bounding boxes from the pseudo-masks, serving as prompts for SAM segment the target image. These methods overlook false positives within the pseudo-mask and exhibit high sensitivity to hyperparameters. Consequently, it heavily relies on the quality of the pseudo-mask and has poor generalization.

Toward this end, we propose a straightforward and effective Visual Reference Prompt (VRP) encoder using meta-learning technique, integrated with SAM to create VRP-SAM. It leverages annotated reference images as prompts to segment similar semantic objects in the target image. As illustrated in Figure~\ref{fig:vs_sam}(b), the VRP encoder accepts inputs in various annotation formats, including points, scribbles, boxes, and masks. Specifically, the VRP encoder introduces a semantic-related model to encode reference and target images into the same space. Following meta-learning methods, we first extract prototypes of target objects from annotated information in reference images, enhancing the representation of the target objects in both images. Following meta-learning methods, the prototypes of target objects (users' markers) are first generated from annotated reference images, which aim to highlight such target instances in both reference and target images.
Then, we introduce a set of learnable queries to extract the semantic cues of the target objects from attentive enhanced reference features. Then, these queries interact with target images, generating prompt embeddings usable by mask decoder to segment semantically specific objects in the target image. Building upon the foundation of SAM's inherent capabilities, VRP-SAM enhances the model's visual reference segmentation prowess. The introduction of Visual Reference Prompts (VRP) not only diversifies the prompts, enabling the model to swiftly segment objects with identical semantics, but also incorporates a meta-learning mechanism that significantly boosts the generalization of model, particularly in dealing with novel objects and cross-domain scenarios. 

To quantitatively assess the generalization capability of VRP-SAM, we follow the dataset configurations commonly used in few-shot segmentation and evaluate our VRP-SAM on Pascal-5$^i$ and COCO-20$i$ datasets. Extensive experiments show that VRP-SAM overcomes the limitations of SAM in prompt format, enabling efficient visual reference segmentation. It is significantly better than sampling-based methods, achieving state-of-the-art results on Pascal-5$^i$ and COCO-20$i$ datasets. Moreover, we present solid evidence of VRP-SAM's superior performance in handling unknown objects and cross-domain scenarios. Our experiments highlight that VRP-SAM achieves the rapid segmentation of a large number of images based on semantics. Furthermore, the incorporation of meta-learning principles significantly enhances its generalization, making it applicable across various scenarios.

\pdfoutput=1
\section{Related Work}
\subsection{Application of SAM}
Segmentation Anything model (SAM)~\cite{sam} is a recently introduced category-agnostic interactive segmentation model by Meta. It leverages user-guided instructions for segmentation, such as points, bounding boxes, and coarse masks. Currently, SAM has two primary application approaches. One involves using SAM's segmentation results as prior information to assist downstream tasks. For instance, Inpaint Anything (IA)~\cite{inpaint} and Edit Everything~\cite{edit} leverage SAM's segmentation results for image editing and restoration~\cite{matte,personalize} within the masked regions. Additionally, SEPL~\cite{SEPL} employs SAM's segmentation results to enhance pseudo-labels in weakly supervised segmentation tasks~\cite{ru2023token,sun2021ssa}. Furthermore, SAM segmentation results serve as prior information in tasks such as crack and volcano crater detection~\cite{ahmadi2023application, giannakis2023deep}. 

The second involves guiding SAM's segmentation through various prompt combinations. For example, in the context of visual reference segmentation~\cite{SVF}, Matcher~\cite{matcher} samples points and boxes from pseudo-masks, utilizing SAM to refine these pseudo-masks. TAM~\cite{yang2023track} is applied in object tracking tasks, where it uses point prompts to initialize object masks and subsequently uses SAM to refine low-quality masks. Additionally, SAMAug~\cite{samaug} introduces a visual point enhancement method tailored for SAM, facilitating automatic image annotation. These examples illustrate SAM's effectiveness across different tasks. However, existing methods are constrained by SAM's existing prompt modalities, and they may struggle when confronted with complex objects and unfamiliar scenes. To break these limitations, we have designed a visual prompt encoder for SAM, expanding its applicability to a wider range of scenarios.

\subsection{Visual reference segmentation}
Visual reference segmentation~\cite{SVF,seem,pfenet} aims to guide the segmentation of a target image using a reference image. The goal of this task is to utilize a semantically annotated reference image to instruct the segmentation of objects or regions in the target image that share the same semantics as those in the reference image. In current research, methods can be broadly categorized into two groups: prototype-based and feature-matching-based. Prototype-based methods, such as PFENet~\cite{pfenet}, PANet~\cite{panet}, and CWT~\cite{simpler}, usually focus on distinguishing prototypes with different class-specific features. ASGNet~\cite{adaptive}, on the other hand, improves the segmentation performance by increasing the number of prototypes. Another feature-matching approach~\cite{vinyals2016matching,hsnet} leverages the pixel-level correlations between reference and target images to significantly enhance segmentation performance, as demonstrated by methods like CyCTR~\cite{cyctr} and HDMNet~\cite{hdmnet}. Moreover, modern large-scale vision models~\cite{seem,painter,bar2022visual} have recognized visual reference segmentation as a primary task, given its indispensable role in handling complex objects and unknown scenes. However, it's important to note that SAM~\cite{sam} does not possess the capability to perform this task, underscoring the necessity of introducing VPR-SAM.
\pdfoutput=1
\begin{figure*}
	\centering
	\includegraphics[width=1.\linewidth]{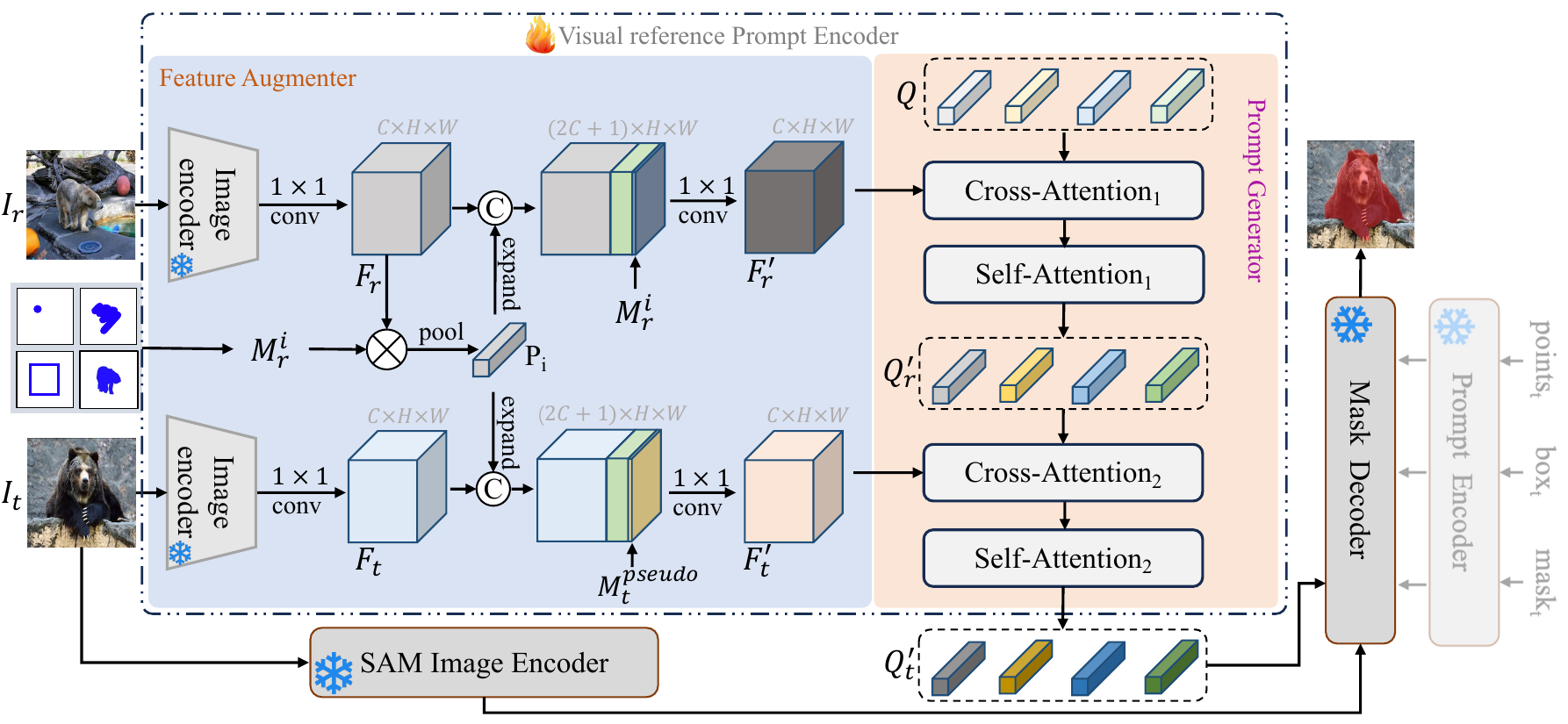}
	\vspace{-1.5em}
	\caption{Proposed VRP-SAM framework. Our approach enables SAM to perform visual reference segmentation by extends a VRP encoder. It takes various granularities of visual references as inputs and encodes these visual references into prompt embeddings. Our VRP encoder consists of a feature augmenter and a prompt generator.}
	\label{fig:structure}
 \vspace{-1.7em}
\end{figure*}

\section{Preliminary}

In this section, we first review the architecture of Segment Anything Model. Then we formulate the problem setting in this paper.

\subsection{SAM Architecture}
SAM is an interactive segmentation model composed of an image encoder, a prompt encoder, and a mask decoder. Given a target image $I_t$ and some geometric prompts, SAM first employs a Vision Transformer (ViT)~\cite{vaswani2017attention} as image encoder to extract image features. Subsequently, the prompt encoder is utilized to generate the prompt embeddings derived from the user-provided points, boxes, or masks. Finally, the mask decoder integrates the image features and prompt embeddings, and generates a high-quality mask in a class-agnostic manner. 

To the best of our knowledge, current research based on SAM still relies on geometric prompts for segmentation. Thus far, there have been no efforts to introduce new forms of prompts to enhance SAM. We design a novel visual reference prompt to extend the visual reference segmentation capabilities of SAM.

\subsection{Problem Formulation}
Visual reference segmentation aims to segment objects in target image that same semantic as the annotated object in reference image. Specifically, let \(I_t\) represent the target image and \(I_r\) represent the reference image. Given a reference image \(I_r\) and its annotation \(M_r^i\), where \(i\) denotes the category of the annotated object,  we leverage this information to segment all regions in the target image belonging to category \(i\).  Depending on the annotation granularity, the reference image have four annotation formats: point, scribble, box, and mask. Point annotation involves providing specific points on the target, scribble requires outlining the target region with arbitrary curves, boxes correspond to bounding boxes around the target, and masks provide pixel-level annotations for the entire target.

\section{VRP-SAM}
VRP-SAM extends SAM to perform visual reference segmentation without compromising its original functionality.
We propose a training-efficient visual reference prompt encoder that, firstly, accommodates various granularities of visual references and, secondly, directly encodes these visual references into prompt embeddings rather than geometric prompts. These prompt embeddings are then fed directly into the mask decoder of SAM, resulting in the generation of the target mask.

As depicted in Figure~\ref{fig:structure}, the VRP Encoder consists of feature augmenter and prompt generator. Next, we will delve into the details of VRP Encoder and the loss function.

\subsection{Feature Augmenter}

Inspired by meta-learning, the Feature Augmenter separately encodes reference annotations \(M_r^i\) into features of both the reference and target images. This process is designed to  distinguish between foreground and background representations. To capture semantic correlations between reference and target images, we introduce a semantic-aware image encoder within the VRP encoder, encoding them into the same latent space.
To prevent overfitting of the VRP encoder, we freeze the image encoder during the training phase. This ensures a balance between capturing semantic relevance and preventing excessive specialization of the VRP encoder.

The Feature Augmenter is illustrated in Figure~\ref{fig:structure}. Initially, leveraging a semantic-aware image encoder (e.g. ResNet-50), we encode \(I_r\) and \(I_t\) separately, resulting in the feature map \(F_r \in \mathbf{R}^{C \times H \times W}\) and \(F_t \in \mathbf{R}^{C \times H \times W}\). Subsequently, we extract the prototype feature \(P_i\) corresponding to class \(i\) from \(F_r\) using the mask \(M_r^i\). This process can be summarized as follows:
\begin{equation}
P_i = MaskAvgPool(F_r, M_r^i)
\end{equation}
where \(M_r^i\) represents one of following annotation formats: \textbf{point}, \textbf{scribble}, \textbf{box}, or \textbf{mask}. To enhance the context information about class \(i\), we concatenate the prototype features and mask with \(F_r\) and \(F_t\). \(F_r\) is concatenated with mask \(m_i\), and \(F_t\) is concatenated with pseudo-mask \(m_{i}^{\text{pseudo}}\). \(m_{i}^{\text{pseudo}}\) is obtained using a common training-free approach, and detailed descriptions are provided in the Appendix. Subsequently, we employ a shared $1\times1$ convolution layer to reduce dimensionality of enhanced features.This process can be summarized as follows:
\begin{align}
F_r^{'} &= Conv(concat(F_r,P_i,m_i))\\
F_t^{'} &= Conv(concat(F_t,P_i,m_i^{\text{pseudo}}))
\end{align}
Ultimately, we obtain enhanced image features $F_r^{'}\in \mathbf{R}^{C \times H \times W}$ and $F_t^{'}\in \mathbf{R}^{C \times H \times W}$, which have enhanced context information for the category $i$. Thus, the enhanced features comprise foreground representations for class \(i\) and background representations for the other classes. Subsequently, we feed the enhanced features into the Prompt Generator to obtain a set of visual reference prompts.

\subsection{Prompt Generator}
The purpose of the Prompt Generator is to obtain a set of visual reference prompts embedding for the SAM mask decoder. As depicted in Figure~\ref{fig:structure}, the process commences with the introduction of a set of learnable queries denoted as \(Q \in \mathbf{R}^{N \times C}\), where \(N\) indicates the number of visual reference prompts.
These queries first engage with the reference features $F_r^{'}$ to obtain the category-specific information through cross-attention and self-attention layer:
\begin{equation}
Q_r^{'} = \text{SelfAttn}_1(\text{CrossAttn}_1(Q, F_r^{'}))
\end{equation}
Here, we obtain a set of queries, \(Q_r^{'}\) , possessing knowledge about the object to be segmented. 
Subsequently, we employ cross-attention to interact these queries with target image feature to obtain the foreground information in target image. Following this, a self-attention layer is used to update the queries, generating a set prompts \(Q_t^{'}\) that align with the representation of SAM : 
\begin{equation}
Q_t^{'} = \text{SelfAttn}_2(\text{CrossAttn}_2(Q_r^{'}, F_t^{'}))
\end{equation}

The final \(Q_t^{'}\) serve as the visual reference prompt embeddings for SAM, equipped with the capability to guide the segmentation of the foreground in the target image. By inputting this set of visual reference prompt embeddings into the mask decoder, the mask \(M_t^{i}\in \mathbf{R}^{1 \times H \times W}\) for category \(i\) in the target image can be obtained.

\subsection{Loss Function}
We employ Binary Cross-Entropy (BCE) loss and Dice loss to supervise the learning of the Visual Reference Prompt Encoder. The BCE loss ensures pixel-wise accuracy, while the Dice loss provides additional context for pixel-level segmentation. Therefore, the total loss of VRP-SAM is:
\vspace{-0.6em}
\begin{align}
L_{\text{total}} &= \underbrace{- \frac{1}{N} \sum_{i=1}^{N} \left[ y_i \log(p_i) + (1 - y_i) \log(1 - p_i) \right]}_{\text{BCE Loss}} \nonumber\\ &+ \underbrace{1 - \frac{2 \sum_{i=1}^{N} (p_i \cdot y_i)}{\sum_{i=1}^{N} p_i^2 + \sum_{i=1}^{N} y_i^2}}_{\text{Dice Loss}} 
\end{align}
where \(N\) represents the total number of pixels, \(y_i\) is the ground truth label for pixel \(i\), and \(p_i\) is the predicted probability of pixel \(i\) belonging to the object. By combining these two losses, we comprehensively consider both accuracy and contextual effects, thus guiding the Visual Reference Prompt Encoder more effectively in generating precise segmentation results.

\begin{table}[!t]
    \caption{Compare with other foundation models. Results of one-shot semantic segmentation on COCO-20$^i$.\textcolor{gray}{Gray} indicates the model is trained by in-domain datasets. $^\dag$ indicates the method using SAM.}
    \vspace{-.6em}
 \label{tab:compare_foundation}
 \centering
 \renewcommand\arraystretch{1.2}
 \setlength{\tabcolsep}{2.2mm}{
 \resizebox{1.0\linewidth}{!}{
\begin{tabular}{c|c|cccc|c}
\hline
\textbf{Methods}                  & \textbf{Label type}            & \textbf{F-0}  & \textbf{F-1}  & \textbf{F-2}  & \textbf{F-3}  & \textbf{Means} \\ \hline
\textcolor{gray}{Painter}~\cite{painter}                  & \multirow{5}{*}{\emph{mask.}} & \textcolor{gray}{31.2} & \textcolor{gray}{35.3} & \textcolor{gray}{33.5} & \textcolor{gray}{32.4} & \textcolor{gray}{33.1}  \\ 
\textcolor{gray}{SegGPT}~\cite{wang2023seggpt}                   &                       & \textcolor{gray}{56.3} & \textcolor{gray}{57.4} & \textcolor{gray}{58.9} & \textcolor{gray}{51.7} & \textcolor{gray}{56.1}  \\  
PerSAM$^\dag$~\cite{persam}                   &                       & 23.1 & 23.6 & 22.0 & 23.4 & 23.0  \\  
PerSAM-F$^\dag$~\cite{persam}                  &                       & 22.3 & 24.0 & 23.4 & 24.1 & 23.5  \\  
Matcher$^\dag$~\cite{matcher}                  &                       & \textcolor{red}{52.7} & \textcolor{blue}{53.5} & 52.6 & \textcolor{red}{52.1} & \textcolor{blue}{52.7}  \\ \hline
\multirow{4}{*}{VRP-SAM$^\dag$} & \emph{point.}                 & 30.1 & 39.2 & 43.0 & 40.4 & 38.2  \\ 
                         & \emph{scribble.}              & \cellcolor{gray!10}40.2 & \cellcolor{gray!10}52.0 & \cellcolor{gray!10}52.4 & \cellcolor{gray!10}44.4 & \cellcolor{gray!10}47.2  \\ 
                         & \emph{box.}                   & \cellcolor{gray!15}44.5 & \cellcolor{gray!15}49.3 & \cellcolor{gray!15}\textcolor{blue}{55.7} & \cellcolor{gray!15}49.1 & \cellcolor{gray!15}49.7  \\ 
                         & \emph{mask.}                  & \cellcolor{gray!25}\textcolor{blue}{48.1} & \cellcolor{gray!25}\textcolor{red}{55.8} & \cellcolor{gray!25}\textcolor{red}{60.0} & \cellcolor{gray!25}\textcolor{blue}{51.6} & \cellcolor{gray!25}\textcolor{red}{53.9}  \\ \hline
\end{tabular}}}
\vspace{-1.5em}
\end{table}
\vspace{-.5em}
\section{Experiments}
\subsection{Setting}

\textbf{Datasets}: To validate segmentation performance and generalization capability of VRP-SAM, we conducted extensive experiments following the few-shot setting~\cite{SVF,cyctr,bam} on COCO-20$^i$~\cite{nguyen2019feature} and PASCAL-5$^i$~\cite{shaban2017one} datasets. Specifically, we organized all classes from both datasets into 4 folds. For each fold, PASCAL-5$^i$~\cite{shaban2017one} comprises 15 base classes for training and 5 novel classes for testing, while COCO-20$^i$~\cite{nguyen2019feature} includes 60 training base classes and 20 testing novel classes. To assess performance of model, we randomly sampled 1000 reference-target pairs in each fold. In each fold, As the mentioned datasets lack labels for point, scribble, and box annotations, we followed the SEEM~\cite{seem} to generate these annotation labels by simulating user inputs randomly based on the reference ground truth masks.\\

\textbf{Implementation details}: In visual reference prompt encoder, we use VGG-16~\cite{vgg} and ResNet-50~\cite{he2016deep} as the image encoder and initialize it with ImageNet~\cite{imagenet} pre-trained weights. We employed the AdamW optimizer~\cite{adamw} along with a cosine learning rate decay strategy for training VRP-SAM. Specifically, on the COCO-20$^i$ dataset, we conducted 50 epochs of training with an initial learning rate of 1e-4 and a batch size of 8. For the PASCAL-5$^i$ dataset, the model was trained for 100 epochs with an initial learning rate of 2e-4 and a batch size of 8. In VRP, the number of queries is set to 50 by default, and the input image size of all experiments needs to be adjusted to $512 \times 512$. Following SEEM~\cite{seem}, during training, we obtain annotations for points, scribbles, and boxes based on mask annotations. We provide detailed descriptions of this process in the Appendix. To ensure a fair comparison, VRP-SAM is exclusively compared to previous works~\cite{hdmnet,painter,li2021ctnet} using visual reference prompts based on the the mean intersection over union (mIoU).

\subsection{Comparison with the State-of-the-art}

\textbf{Comparison with other foundation models.} Leveraging the foundation model enables few-shot segmentation. We compared various approaches utilizing Painter~\cite{painter}, SegGPT~\cite{wang2023seggpt}, and SAM as foundation models for few-shot segmentation, providing evaluation metrics on the COCO-20$^i$ dataset. As shown in Table~\ref{tab:compare_foundation}, VRP-SAM achieves 53.9\% mean mIoU without training on novel classes, achieving comparable with SegGPT. Note that the training data of SegGPT include all classes in COCO. Furthermore, our VRP-SAM outperforms other SAM-base methods including Matcher~\cite{matcher}, which employs a DINOv2~\cite{dinov2} pretrained ViT-L~\cite{vaswani2017attention} as image encoder.

\begin{table*}[!t]\footnotesize
    \caption{Performance of one-shot semantic segmentation on COCO-20$^i$ and PASCAL-5$^i$. The \textcolor{red}{red} and \textcolor{blue}{blue} colors respectively represent the optimal and suboptimal results.}
    \vspace{-.5em}
 \label{tab:coco_mask}
 \centering
 \renewcommand\arraystretch{1.2}
 \setlength{\tabcolsep}{3.4mm}{
 \resizebox{1.0\linewidth}{!}{
\begin{tabular}{c|c|c|ccccc|ccccc}
\hline
\multirow{2}{*}{\textbf{Method}} &
  \multirow{2}{*}{\begin{tabular}[c]{@{}c@{}}\textbf{Image} \\ \textbf{encoder}\end{tabular}} &
  \multirow{2}{*}{\begin{tabular}[c]{@{}c@{}}\textbf{Learnable}\\  \textbf{params}\end{tabular}} &
  \multicolumn{5}{c|}{\textbf{COCO-20$^i$}} &
  \multicolumn{5}{c}{\textbf{PASCAL-5$^i$}} \\ \cline{4-13} 
 && & F-0 & F-1 & F-2 & \multicolumn{1}{c|}{F-3} & Mean & F-0 & F-1 & F-2 & \multicolumn{1}{c|}{F-3} &  Mean \\ \hline
PFENet~\cite{pfenet} & \multirow{4}{*}{VGG-16} & 10.4M & 35.4 & 38.1 & 36.8 & \multicolumn{1}{c|}{34.7} & 36.3 & 56.9 & 68.2 & 54.5 & \multicolumn{1}{c|}{52.4} & 58.0 \\
BAM~\cite{bam} &  & 4.9M & 36.4 & 47.1 & 43.3 & \multicolumn{1}{c|}{41.7} & 42.1 & 63.2 & 70.8 & 66.1 & \multicolumn{1}{c|}{57.5} & 64.4 \\
HDMNet~\cite{hdmnet} &  &4.2M & 40.7 &50.6 &48.2 &\multicolumn{1}{c|}{44.0} &45.9 &64.8 &71.4 &67.7 &\multicolumn{1}{c|}{56.4} &65.1 \\
\rowcolor{gray!25}
 VRP-SAM & &1.6M &43.6 &51.7 &50.0 &\multicolumn{1}{c|}{46.5} &48.0 &70.0 &74.7 &68.3 &\multicolumn{1}{c|}{61.9} &68.7 \\ \hline
PFENet~\cite{pfenet} &\multirow{8}{*}{ResNet-50} &10.4M &36.5 &38.6 &34.5 &\multicolumn{1}{c|}{33.8} &35.8 & 61.7 &69.5 & 55.4 &\multicolumn{1}{c|}{56.3} & 60.8 \\
HSNet~\cite{hsnet} & &2.6M &36.3 & 43.1 & 38.7 &\multicolumn{1}{c|}{38.7} &39.2 &64.3 &70.7 &60.3 &\multicolumn{1}{c|}{60.5} & 64.0 \\
CyCTR~\cite{cyctr} & & 15.4M & 38.9 &43.0 &39.6 &\multicolumn{1}{c|}{39.8} &40.3 &65.7 &71.0 &59.5 &\multicolumn{1}{c|}{59.7} & 64.0 \\
SSP~\cite{ssp} & & 8.7M & 35.5 &39.6 &37.9 &\multicolumn{1}{c|}{36.7} &37.4 &60.5 &67.8 &66.4 &\multicolumn{1}{c|}{51.0} & 61.4 \\
NTRENet~\cite{ntre} & & 19.9M & 36.8 &42.6 &39.9 &\multicolumn{1}{c|}{37.9} &39.3 &65.4 &72.3 &59.4 &\multicolumn{1}{c|}{59.8} & 64.2 \\
DPCN~\cite{dpcn}& & - & 42.0 &47.0 &43.3 &\multicolumn{1}{c|}{39.7} &43.0 &65.7 &71.6 &69.1 &\multicolumn{1}{c|}{60.6} & 66.7 \\
VAT~\cite{vat} & & 3.2M & 39.0 &43.8 &42.6 &\multicolumn{1}{c|}{39.7} &41.3 &67.6 &72.0 &62.3 &\multicolumn{1}{c|}{60.1} & 65.5 \\

BAM~\cite{bam} & &4.9M & 39.4 & 49.9 & 46.2 &\multicolumn{1}{c|}{45.2} &45.2 &69.0 &73.6 & 67.6 & \multicolumn{1}{c|}{61.1} & 67.8 \\
HDMNet~\cite{hdmnet} & &4.2M & 43.8 & \textcolor{blue}{55.3} & 51.6 &\multicolumn{1}{c|}{\textcolor{blue}{49.4}} &50.0 &71.0 & \textcolor{blue}{75.4} & \textcolor{blue}{68.9} &\multicolumn{1}{c|}{62.1} &\textcolor{blue}{69.4} \\
\rowcolor{gray!25}
VRP-SAM & & 1.6M &\textcolor{blue}{48.1} &\textcolor{red}{55.8} &\textcolor{red}{60.0} &\multicolumn{1}{c|}{\textcolor{red}{51.6}} &\textcolor{red}{53.9} &\textcolor{red}{73.9} &\textcolor{red}{78.3} &\textcolor{red}{70.6} &\multicolumn{1}{c|}{\textcolor{blue}{65.0}} &\textcolor{red}{71.9} \\ \hline
DCAMA &Swin-B &47.7M & \textcolor{red}{49.5} &52.7 &\textcolor{blue}{52.8} &\multicolumn{1}{c|}{48.7} &50.9 &72.2 &73.8 &64.3 &\multicolumn{1}{c|}{\textcolor{red}{67.1}} &69.3 \\ \hline
\end{tabular}}}
\vspace{-1em}
\end{table*}

\textbf{Comparison with few-shot methods.} To validate the effectiveness of VRP-SAM, we compared it with state-of-the-art few-shot segmentors on the novel set of COCO-20$^i$ and PASCAL-5$^i$ datasets. To ensure a fair comparison, VRP-SAM is trained and tested separately in each fold, ensuring no overlap in classes between the training and test sets. The results in Table~\ref{tab:coco_mask} demonstrate that VRP-SAM achieves state-of-the-art results on COCO-20$^i$ and PASCAL-5$^i$, with mIoU scores of 53.9 and 71.9, respectively. Notably, when using VGG-16 as the image encoder for VRP, VRP-SAM achieves mIoU scores of 48.0 and 68.7 on COCO-20$^i$ and PASCAL-5$^i$ datasets. It indicates that VRP-SAM achieves optimal performance on the novel set with only 1.6M learnable parameters, highlighting its powerful generalization capability. \\

\begin{table}[]\footnotesize
    \caption{Comparison with geometric prompts. Geometric prompts are randomly sampled from the pseudo-mask. $^\dag$ indicates the carefully designed sampling strategy proposed in~\cite{matcher}.}
 \label{tab:comp_sam}
 \centering
 \renewcommand\arraystretch{1.2}
\begin{tabular}{c|c|c|c}
\hline
\textbf{Method} & \textbf{Image Encoder} & \textbf{Prompts} & \textbf{Mean IoU} \\
\hline
\multirow{7}{*}{GP-SAM}    & \multirow{3}{*}{ResNet-50}  & \emph{box.}         &  19.7     \\ 
        &               & \emph{point.}       &  21.7     \\ 
        &               & \emph{box.} + \emph{point.} & 23.2      \\ \cline{2-4}
     & \multirow{4}{*}{DINOv2}       & \emph{box.}         &  31.3     \\ 
        &               & \emph{point.}       & 36.6     \\ 
        &               & \emph{box.} + \emph{point.} & 37.8     \\
        &               & \emph{box.} + \emph{point.} $^{\dag}$~\cite{matcher}   & 52.7 \\ \hline\hline
\multirow{5}{*}{VRP-SAM} & \multirow{4}{*}{ResNet-50}      & \emph{point.}       & 38.4      \\ 
        &               &  \cellcolor{gray!10}\emph{scribble.}    & 47.3      \\ 
        &               &  \cellcolor{gray!15}\emph{box.}         & 49.7     \\ 
        &               &  \cellcolor{gray!25}\emph{mask.}        & \textcolor{blue}{53.9}     \\ \hline
& DINOv2-B/14 &\cellcolor{gray!25}\emph{mask.} & \textcolor{red}{60.4}\\ \hline
\end{tabular}
\vspace{-1.5em}
\end{table}

\vspace{-1.5em}
\subsection{Comparison with Geometric Prompts}
\begin{figure}[t]
\centering
\includegraphics[clip, trim=80 40 50 50, width=0.476\textwidth]{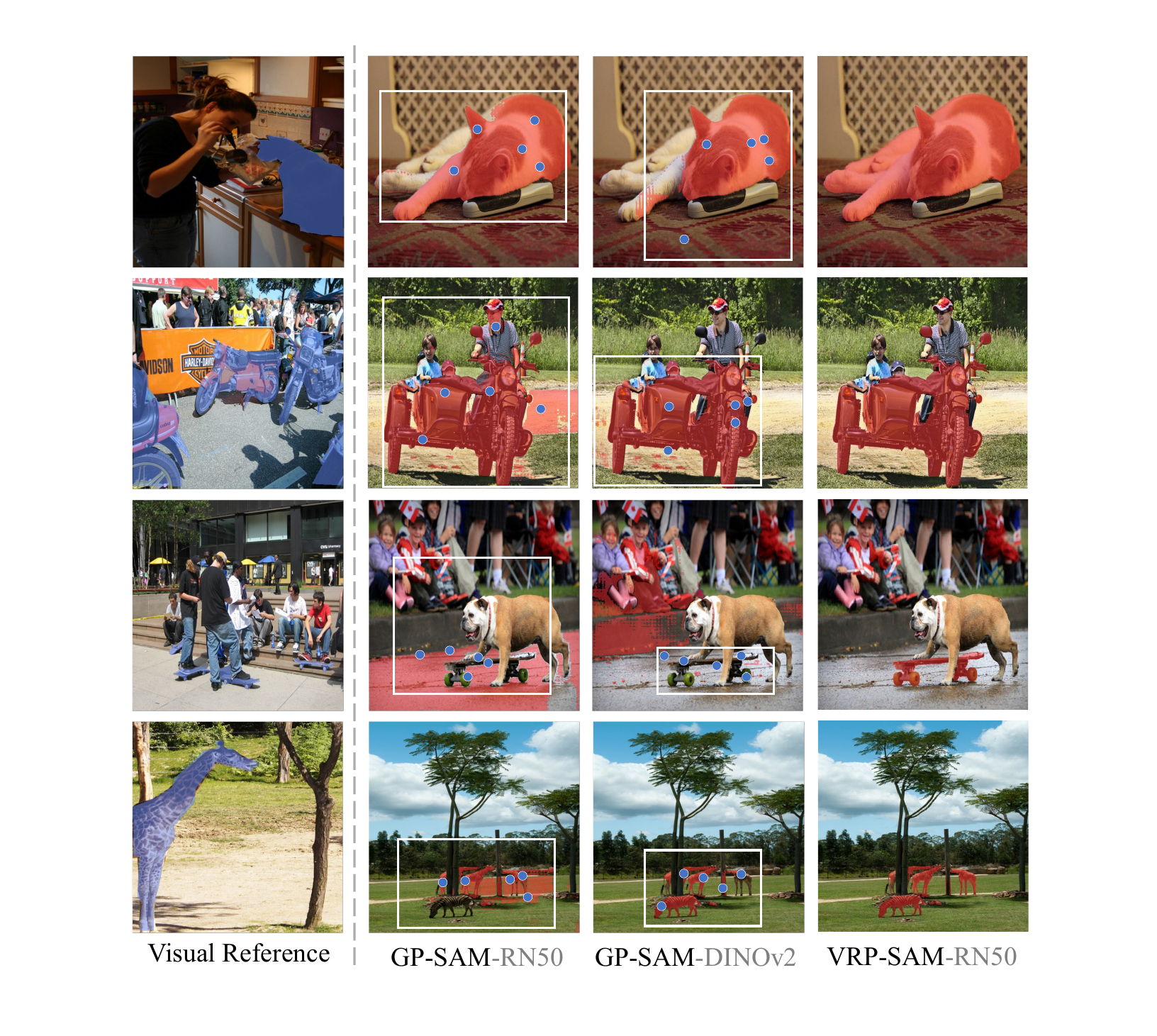} 
\vspace{-1.5em}
\caption{The visualization results of VRP-SAM and GP-SAM.}
\label{fig:vis_coco}
\vspace{-2.em}
\end{figure}

In this paper, we design Visual Reference Prompts to represent target information. Different from the Geometric Prompts (GP) provided by SAM, our VRP is more flexible and robust. We conducted experiments comparing GP and VRP to validate the superiority of our method (see Table~\ref{tab:comp_sam}).
In the GP-SAM experiments, we initially used an Image Encoder to generate a pseudo-mask for the target image following~\cite{pfenet}, and subsequently obtained points or bounding boxes from the pseudo-mask~\footnote{For point prompts, we randomly sample 5 points from the pseudo-mask. For box prompt, we adopt the bounding box of the pseudo-mask.} as geometric prompts. Experimental results consistently demonstrate the ongoing superiority of our VRP over the GP approach. We visualize the segmentation results of GP-SAM and our VRP-SAM in Figure~\ref{fig:vis_coco}. Visualization results indicate that the GP approach is prone to generating false-positive prompts, significantly impacting segmentation performance. In contrast, our VRP effectively avoids such issues.

\begin{figure*}
	\centering
	\includegraphics[width=\linewidth]{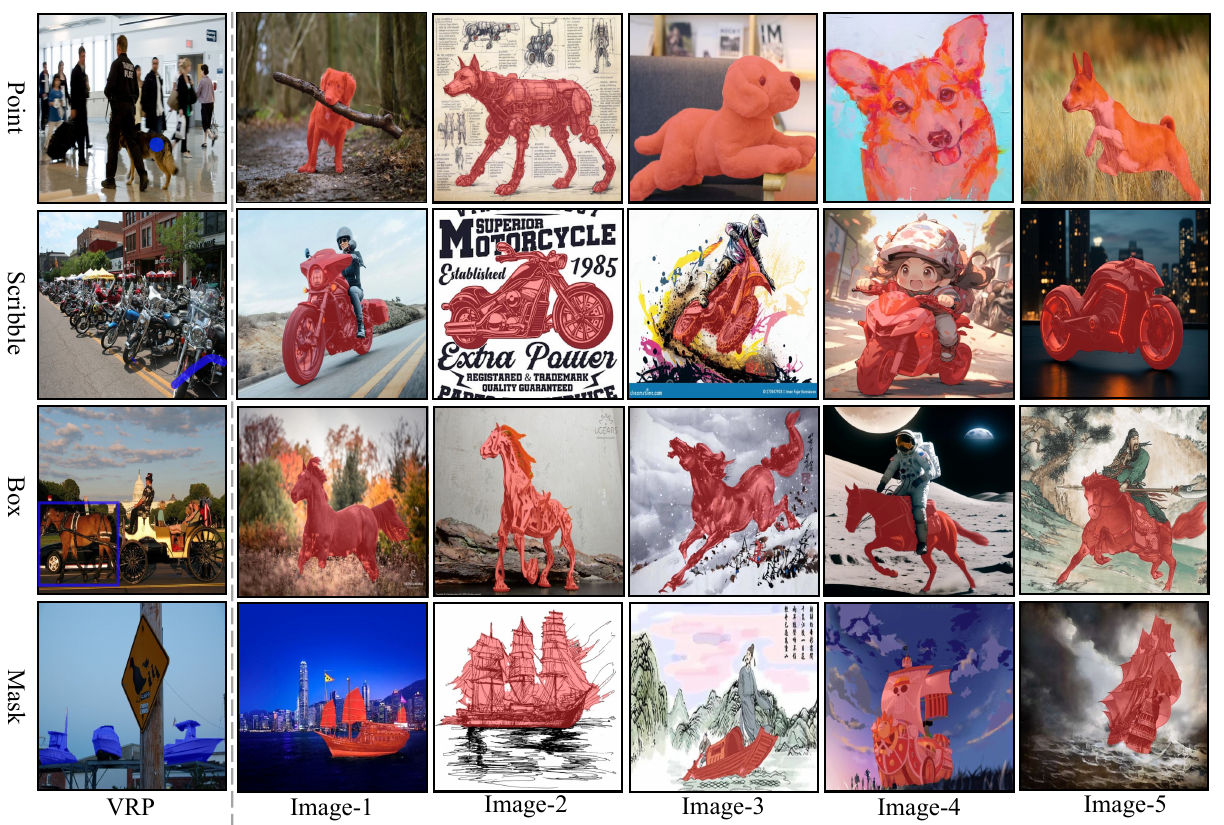}
	\vspace{-1.5em}
	\caption{Qualitative results of VRP-SAM across diverse image styles is presented. The target images were sourced from the internet.}
	\label{fig:web}
 \vspace{-1.5em}
\end{figure*}
\vspace{-1.5mm}
\subsection{Generalization Evaluation}

\textbf{Domain shift:} Next, we assess the effectiveness of VRP-SAM in a domain shift scenario, which necessitates significant domain differences between training and testing sets. Following prior work~\cite{hsnet,pfenet,sun2023exploring}, we trained on COCO-20$^i$ and tested on PASCAL-5$^i$, where the classes in training set do not overlap with those in test set. The results in Table~\ref{tab:coco2PASCAL}, representing the average across four folds, clearly demonstrate the superior performance of VRP-SAM in domain shift scenario. Notably, VRP-SAM with scribble annotations outperforms FP-Trans~\cite{zhang2022feature} by 2.8\%, while VRP-SAM with mask annotations surpasses DGPNet~\cite{johnander2022dense} by 5.8\%. This affirms the robust generalization capability of VRP-SAM and its effectiveness in domain transfer scenarios.\\
\textbf{Visualization:} To assess the generalization capability of VRP-SAM across diverse image styles, we curated a collection of target images from web, spanning various genres such as natural landscapes, artworks, and complex environments. Visual Reference Prompts (VRPs) were selected from the COCO dataset to guide the segmentation of these target images. The segmentation results on different image styles are presented in Figure~\ref{fig:web}. It demonstrate that VRP-SAM adeptly adapts to varied image styles, accurately delineating target objects. Notably, VRP-SAM excels in handling ink-style images, showcasing refined segmentation performance. This compellingly underscores the robustness and effectiveness of VRP-SAM when confronted with images of unknown styles.

\begin{table}[!t]\footnotesize
    \caption{Evaluation (Mean IoU (\%)) under the
domain shift from COCO-20$^i$ to PASCAL-5$^i$.}
\vspace{-.5em}
 \label{tab:coco2PASCAL}
 \centering
 \renewcommand\arraystretch{1.1}
 \setlength{\tabcolsep}{3.4mm}{
 \resizebox{1.0\linewidth}{!}{
\begin{tabular}{c|c|c|c}
\hline
\textbf{Method}     & \textbf{Image encoder}         & \textbf{Label type}            & \textbf{Mean} \\ \hline
RPMM~\cite{yang2020prototype}   & \multirow{4}{*}{ResNet-50}  & \multirow{4}{*}{\emph{mask.}} & 49.6 \\  
PFENet~\cite{pfenet}  &                        &                       & 61.1 \\  
RePRI~\cite{boudiaf2021few}    &                        &                       & 63.2 \\  
VAT-HM~\cite{moon2022hm}   &                        &                       & 65.1 \\ \hline
HSNet~\cite{hsnet}   & \multirow{2}{*}{ResNet-101} & \multirow{2}{*}{\emph{mask.}}                  & 64.1 \\ 
DGPNet~\cite{johnander2022dense}   &                        &                  & 70.1 \\ \hline
FP-Trans~\cite{zhang2022feature} & DeiT-B/16              & \emph{mask.}                  & 69.7 \\ \hline \hline
\multirow{4}{*}{VRP-SAM}  & \multirow{4}{*}{ResNet-50}  & \emph{point.}                 & 63.5 \\ 

  &    & \cellcolor{gray!10} \emph{scribble.}  & \cellcolor{gray!10} \textcolor{blue}{72.5} \\ 

 &       & \cellcolor{gray!15} \emph{box.}    & \cellcolor{gray!15} 72.3 \\ 
 &    & \cellcolor{gray!25} \emph{mask.}    & \cellcolor{gray!25} \textcolor{red}{75.9} \\ \hline
\end{tabular}}}
\vspace{-2.8em}
\end{table}
\vspace{-.5em}
\subsection{Ablation Study}

\begin{figure}[t]
\centering
\includegraphics[width=0.47\textwidth]{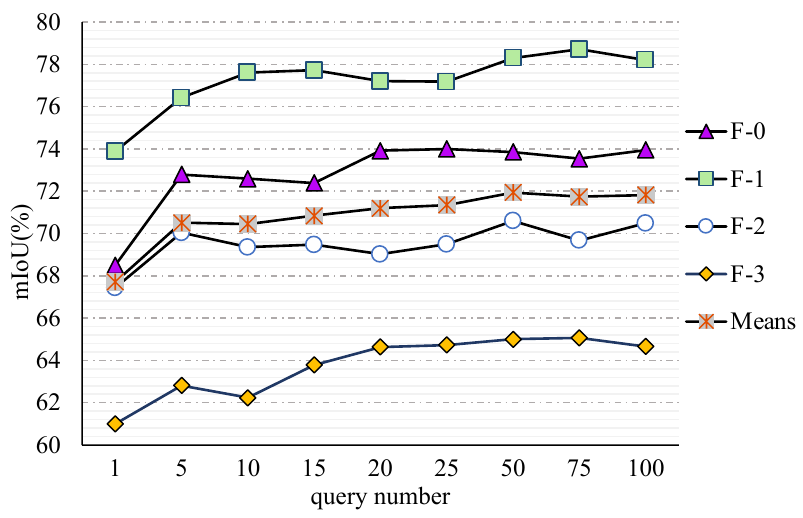} 
\vspace{-1.em}
\caption{Ablation study on VRP-SAM with different query numbers. The x-axis shows the number of queries, and the y-axis represents model performance. }
\label{fig:query}
\vspace{-2em}
\end{figure}

To validate the effectiveness of VRP-SAM, we conducted extensive ablation studies on PASCAL-5$^i$. ResNet-50 was chosen as the image encoder for VRP to ensure experimental consistency. These experiments aimed to thoroughly investigate the performance of VRP-SAM under various conditions.\\
\textbf{Loss:} To assess the impact of Binary Cross-Entropy (BCE) and Dice losses on VRP-SAM, experiments were conducted on the PASCAL-5$^i$ dataset. Results in Table~\ref{tab:loss} indicate comparable performance when using BCE or Dice loss individually. Notably, the optimal performance for VRP-SAM is achieved when both losses are combined. It emphasizes the importance of BCE and Dice losses in guiding VRP-SAM to produce more robust and accurate masks. BCE loss ensures precise pixel-level classification, while Dice loss facilitates accurate spatial localization of objects. Simultaneously employing BCE and Dice losses in VRP-SAM maximizes their complementary advantages.\\
\vspace{-.3em}
\textbf{The number of query:} In Figure~\ref{fig:query}, we conducted a comprehensive analysis of the impact of varying query quantities on VRP-SAM's performance. We observed a positive correlation between increased query numbers and improved segmentation quality. However, once the query count surpassed 50, the performance gain started to diminish. This phenomenon suggests that 50 queries provide ample effective guidance for the model, and further increases in query count do not yield significant improvements, leading to performance saturation. To maintain high performance while minimizing model learnable parameters, we set the query quantity in VRP-SAM to 50. This decision optimally balances guidance information and model efficiency.\\
\textbf{Initialization of query:} In Table~\ref{tab:init}, we compared various query initialization strategies, such as random initialization, foreground prototype (FP), background prototype (BP), and a hybrid prototype with half foreground and half background (half-FP \& half-BP). Surprisingly, random initialization outperformed all other strategies, showcasing superior performance. This unexpected outcome can be attributed to the intricate segmentation task and the dataset's object diversity. Random initialization enables the model to explore a broader range of states, avoiding potential local minima. In contrast, prototype-based initializations, whether foreground, background, or a combination, might introduce biases, limiting adaptability to diverse object characteristics.\\

\begin{table}[]
    \caption{Ablation study on different loss function in VRP-SAM.}
\vspace{-.5em}
 \label{tab:loss}
 \centering
 \renewcommand\arraystretch{1.2}
 \setlength{\tabcolsep}{3.4mm}{
 \resizebox{1.0\linewidth}{!}{
    \begin{tabular}{c|c|cc|c}
    \hline
        \textbf{Method} & \textbf{Label type} & \textbf{Bce loss} & \textbf{Dice loss} & \textbf{Means} \\ \hline
        \multirow{3}{*}{VRP-SAM} & \multirow{3}{*}{\emph{mask.}} &  \usym{2713} & \usym{2717} & 70.1 \\ 
         &  & \usym{2717} & \usym{2713} & 70.3 \\ 
         &  & \cellcolor{gray!25} \usym{2713} & \cellcolor{gray!25} \usym{2713} & \cellcolor{gray!25} \textbf{71.9} \\ \hline
    \end{tabular}}}
    \vspace{-1.3em}
\end{table}
\vspace{-1.5em}
\textbf{Number of VRPs:} In our investigation of the influence of number of visual reference prompts on segmentation results, we explored utilization of few visual reference prompts. This approach involves sending several visual reference prompts to the VRP encoder, generating multiple prompt embeddings. These embeddings are concatenated and forwarded to the mask decoder, resulting in the final mask. The result in Table~\ref{tab:number}, we observed a substantial enhancement in segmentation performance with an increase in the number of visual reference prompts. The results emphasize positive influence of incorporating multiple visual reference prompts, enhancing the model's ability to accurately capture intricate details and nuances associated with target object. The diverse prompts contribute enriched contextual information, facilitating a more comprehensive understanding of the object's characteristics and leading to generation of precise and nuanced segmentation masks.

\begin{table}[]\footnotesize
    \caption{Ablation study of different query initialization methods on VRP-SAM.}
    \vspace{-.8em}
 \label{tab:init}
 \centering
 \renewcommand\arraystretch{1.2}
    \begin{tabular}{c|c|c|c}
    \hline
        \textbf{Method} & \textbf{Label type} & \textbf{Image encoder} & \textbf{Mean IoU} \\ \hline
        \multirow{4}{*}{VRP-SAM} & \multirow{4}{*}{\emph{mask.}} & FP & 68.2 \\ 
         & &  BP & 62.6 \\
         & & half-FP \& half-BP & 67.4 \\
         & & \cellcolor{gray!25} random & \cellcolor{gray!25} \textbf{71.9} \\ \hline
    \end{tabular}
\vspace{-1.em}
\end{table}

\begin{table}[]\footnotesize
    \caption{Ablation Study on Few Visual Reference prompts for VRP-SAM.}
    \vspace{-.8em}
 \label{tab:number}
 \centering
 \renewcommand\arraystretch{1.2}
 \setlength{\tabcolsep}{3.4mm}{
 \resizebox{1.0\linewidth}{!}{
\begin{tabular}{c|c|cc}
\hline
\textbf{Method} &\textbf{Label type}  & \textbf{1-VRP} & \textbf{5-VRP} \\  \hline
\multirow{4}{*}{VRP-SAM} & \emph{point.}     & \cellcolor{gray!10} 62.9   & \cellcolor{gray!15} 64.1$_{\textcolor{red}{(+1.2)}}$   \\  
                         & \emph{scribble.}  &\cellcolor{gray!10} 66.8   & \cellcolor{gray!15} 68.4$_{\textcolor{red}{(+1.6)}}$       \\ 
                         & \emph{box.}         & \cellcolor{gray!10} 69.4   & \cellcolor{gray!15} 70.5$_{\textcolor{red}{(+1.1)}}$       \\ 
                         & \emph{mask.}      & \cellcolor{gray!10} 71.9   & \cellcolor{gray!15} 72.9$_{\textcolor{red}{(+1.0)}}$       \\ \hline
\end{tabular}}}
\vspace{-2.3em}
\end{table}

\section{Conclusion}

In this paper, we present VRP-SAM, an innovative extension of the SAM framework achieved through integration of a Visual Reference Prompt (VRP) encoder. This addition empowers SAM to leverage visual reference prompts for guided segmentation. The core methodology involves encoding annotated reference images through the VRP encoder, which then interacts with the target image to generate meaningful prompts for segmentation within the SAM framework. The seamless fusion of VRP encoder with SAM, resulting in VRP-SAM, enhances the model's generality and adaptability. VRP-SAM overcomes limitations posed by SAM's existing prompt formats, especially in complex scenarios and large datasets. The introduced visual reference prompts, including point, box, scribble, and mask annotations, offer a flexible solution, expanding the model's applicability. Extensive empirical studies conducted showcase VRP-SAM's state-of-the-art performance in visual reference segmentation with minimal learnable parameters. Notably, VRP-SAM demonstrates robust generalization capabilities, excelling in segmentation tasks for novel objects and cross-domain scenarios. \\
\textbf{Acknowledge} This work was partially supported by the National Key Research and Development Program of China under Grant 2022ZD0118802 and the National Natural Science Foundation of China (Grant No. U20B2064 and U21B2043).
\vspace{+1.em}
\appendix
\noindent
{\Large {\textbf{Appendix}}}
\begin{table}[]\small
    \caption{Details of the data split for PASCAL-5$^i$. Each row consists of 5 classes designated for testing, with the remaining 15 classes utilized for training.}
 \label{tab:data_PASCAL}
 \centering
 \renewcommand\arraystretch{1.3}
 \setlength{\tabcolsep}{6.4mm}{
 \resizebox{1.0\linewidth}{!}{
\begin{tabular}{c|l}
\hline
\textbf{Fold} & \textbf{Test classes}                                 \\ \hline
\textbf{0}    & Aeroplane, Bicycle, Bird, Boat, Bottle       \\ \hline
\textbf{1}    & Bus, Car, Cat, Chair, Cow                    \\ \hline
\textbf{2}    & Dining table, Dog, Horse, Motorbike, Person  \\ \hline
\textbf{3}    & Potted plant, Sheep, Sofa, Train, TV/monitor \\ \hline
\end{tabular}}}
\end{table}

\begin{table}[]
    \caption{Details of the data split for COCO-20$^i$. Each row consists of 20 classes designated for testing, with the remaining 60 classes utilized for training.}
 \label{tab:data_coco}
 \centering
 \renewcommand\arraystretch{1.3}
 \setlength{\tabcolsep}{2.4mm}{
 \resizebox{1.0\linewidth}{!}{
\begin{tabular}{c|l}
\hline
\textbf{Fold} &
  \textbf{Test classes} \\ \hline
\textbf{0} &
  \begin{tabular}[c]{@{}l@{}}Person, Airplane, Boat, Parking meter, Dog, Elephant, Backpack, \\ Suitcase, Sports ball, Skateboard, Wine glass, Spoon, Sandwich, \\ Hot dog, Chair, Dining table, Mouse, Microwave, Sink, Scissors\end{tabular} \\ \hline
\textbf{1} &
  \begin{tabular}[c]{@{}l@{}}Bicycle, Bus, Traffic light, Bench, Horse, Bear, Umbrella, Frisbee, \\ Kite, Surfboard, Cup, Bowl, Spoon, Orange, Pizza, Couch, Toilet, \\ Remote, Oven, Book, Teddy bear\end{tabular} \\ \hline
\textbf{2} &
  \begin{tabular}[c]{@{}l@{}}Car, Train, Fire hydrant, Bird, Sheep, Zebra, Handbag, Skis, \\ Baseball bat, Tennis racket, Fork, Banana, Broccoli, Donut, Potted \\ plant, TV, Keyboard, Toaster, Clock, Hair drier\end{tabular} \\ \hline
\textbf{3} &
  \begin{tabular}[c]{@{}l@{}}Motorcycle, Truck, Stop sign, Cat, Cow, Giraffe, Tie, Snowboard, \\ Baseball glove, Bottle, Knife, Apple, Carrot, Cake, Bed, Laptop, \\ Cell phone, Sink, Vase, Toothbrush\end{tabular} \\ \hline
\end{tabular}}}
\end{table}

\section{Additional Implementation Details}
\textbf{Training:}To capture semantic correlations between reference and target images, we introduced a semantic relevance model in the Visual Prompt Encoder. Specifically, we validated the effectiveness of VRP-SAM on VGG-16~\cite{vgg}, ResNet-50~\cite{he2016deep}, and DINOv2~\cite{dinov2}. Following prior work~\cite{pfenet,SVF}, we utilized mid-level features from the image encoder to retain finer details, with high-level features used to generate a pseudo mask for the target image. For instance, in the case of ResNet-50, mid-level features corresponded to the $3^th$ and $4^th$ blocks, while high-level features were extracted from the $5^th$ block.

\begin{table}[]\small
    \caption{Details of the data split for PASCAL-5$^i$ in domain shift scenario. Each line represents non-overlapping classes in the training set corresponding to the respective fold of COCO.}
 \label{tab:data_coco2PASCAL}
 \centering
 \renewcommand\arraystretch{1.3}
 \setlength{\tabcolsep}{6.4mm}{
 \resizebox{1.0\linewidth}{!}{
\begin{tabular}{c|l}
\hline
\textbf{Fold} & \textbf{Test classes}                                      \\ \hline
\textbf{0}    & Aeroplane, Boat, Chair, Dining table, Dog, Person \\ \hline
\textbf{1}    & Horse, Sofa, Bicycle, Bus                         \\ \hline
\textbf{2}    & Bird, Car, Potted plant, Sheep, Train, TV/monitor \\ \hline
\textbf{3}    & Bottle, Cow, Cat, Motorbike                       \\ \hline
\end{tabular}}}
\vspace{-1.5em}
\end{table}

Within the feature augmenter of the Visual Reference Encoder, the pseudo mask for the target was computed by evaluating the pixel-wise similarity map through the comparison of high-level features of reference and target images. We retained the maximum similarity at each pixel and normalized the similarity map to the [0, 1] range using min-max normalization. This similarity map serves as the pseudo mask for the target image. We followed the SEEM~\cite{seem} approach, where point reference prompts involve randomly sampling $(1, 20)$ points from the ground truth (GT), scribble reference prompts are generated randomly using a free-form training mask generation algorithm proposed in \cite{yu2019free}, resulting in $(1, 20)$ scribbles, and box reference prompts are obtained by extracting object bounding boxes from the GT.\\

\begin{table*}[]\footnotesize
    \caption{Compare with the State-of-the-arts. Results of one-shot semantic segmentation on COCO-20$^i$.Gray indicates the model is trained by in-domain datasets. The \textcolor{red}{red} colors respectively represent the optimal results.}
 \label{tab:app_sota}
 \centering
 \renewcommand\arraystretch{1.3}
 \setlength{\tabcolsep}{5.4mm}{
 \resizebox{1.0\linewidth}{!}{
\begin{tabular}{cccccccc}
\hline
\multicolumn{1}{c|}{\textbf{Method}} & \multicolumn{1}{c|}{\textbf{Venue}} & \multicolumn{1}{c|}{\textbf{Image encoder}} & \textbf{F-0} & \textbf{F-1} & \textbf{F-2} & \multicolumn{1}{c|}{\textbf{F-3}} & \textbf{Mean} \\ \hline
\multicolumn{8}{l}{\textit{few-shot method.}} \\ \hline
\multicolumn{1}{c|}{BAM+SVF~\cite{SVF}} & \multicolumn{1}{c|}{NeurIPS'22} & \multicolumn{1}{c|}{\multirow{3}{*}{ResNet-50}} & 46.9 & 53.8 & 48.4 & \multicolumn{1}{c|}{44.8} & 48.5 \\
\multicolumn{1}{c|}{VAT~\cite{vat}} & \multicolumn{1}{c|}{ECCV'23} & \multicolumn{1}{c|}{} & 39.0 & 43.8 & 42.6 & \multicolumn{1}{c|}{39.7} & 41.3 \\
\multicolumn{1}{c|}{HDMNet~\cite{hdmnet}} & \multicolumn{1}{c|}{CVPR'23} & \multicolumn{1}{c|}{} & 43.8 & 55.3 & 51.6 & \multicolumn{1}{c|}{49.4} & 50.0 \\ \hline
\multicolumn{1}{c|}{FPTrans~\cite{zhang2022feature}} & \multicolumn{1}{c|}{NeurIPS'22} & \multicolumn{1}{c|}{Deit-B/16} & 44.4 & 48.9 & 50.6 & \multicolumn{1}{c|}{44.0} & 47.0 \\ \hline
\multicolumn{1}{c|}{DCAMA~\cite{DCAMA}} & \multicolumn{1}{c|}{ECCV'23} & \multicolumn{1}{c|}{Swin-B} & 49.5 & 52.7 & 52.8 & \multicolumn{1}{c|}{48.7} & 50.9 \\ \hline
\multicolumn{8}{l}{\textit{based foundation methods.}} \\ \hline
\multicolumn{1}{c|}{\textcolor{gray}{Painter~\cite{painter}}} & \multicolumn{1}{c|}{\textcolor{gray}{CVPR'23}} & \multicolumn{1}{c|}{\textcolor{gray}{Painter}} & \textcolor{gray}{31.2} & \textcolor{gray}{35.3} & \textcolor{gray}{33.5} & \multicolumn{1}{c|}{\textcolor{gray}{32.4}} & \textcolor{gray}{33.1} \\ \hline
\multicolumn{1}{c|}{PerSAM~\cite{persam}} & \multicolumn{1}{c|}{arXiv'23} & \multicolumn{1}{c|}{-} & 23.1 & 23.6 & 22.0 & \multicolumn{1}{c|}{23.4} & 23.0 \\ \hline
\multicolumn{1}{c|}{Matcher~\cite{matcher}} & \multicolumn{1}{c|}{arXiv'23} & \multicolumn{1}{c|}{DINOv2-L} & 52.7 & 53.5 & 52.6 & \multicolumn{1}{c|}{52.1} & 52.7 \\ \hline
\multicolumn{1}{c|}{\textcolor{gray}{SegGPT~\cite{wang2023seggpt}}} & \multicolumn{1}{c|}{\textcolor{gray}{ICCV'23}} & \multicolumn{1}{c|}{\textcolor{gray}{Painter}} & \textcolor{gray}{56.3} & \textcolor{gray}{57.4} & \textcolor{gray}{58.9} & \multicolumn{1}{c|}{\textcolor{gray}{51.7}} & \textcolor{gray}{56.1} \\ \hline
\multicolumn{1}{c|}{\multirow{3}{*}{VRP-SAM}} & \multicolumn{1}{c|}{\multirow{3}{*}{This work}} & \multicolumn{1}{c|}{VGG-16} & 43.6 & 51.7 & 50.0 & \multicolumn{1}{c|}{46.5} & 48.0 \\ 
\multicolumn{1}{c|}{} & \multicolumn{1}{c|}{} & \multicolumn{1}{c|}{ResNet-50} & 48.1 & 55.8 & 60.0 & \multicolumn{1}{c|}{51.6} & 53.9 \\ 
\multicolumn{1}{c|}{} & \multicolumn{1}{c|}{} & \multicolumn{1}{c|}{\cellcolor{gray!25} DINOv2-B} & \cellcolor{gray!25} \textcolor{red}{56.8} & \cellcolor{gray!25} \textcolor{red}{61.0} & \cellcolor{gray!25} \textcolor{red}{64.2} & \multicolumn{1}{c|}{\cellcolor{gray!25} \textcolor{red}{59.7}} & \cellcolor{gray!25} \textcolor{red}{60.4} \\ \hline
\end{tabular}}}
\end{table*}

\begin{table*}[]\scriptsize
    \caption{Ablation study based different label types on COCO-20$^i$. The \textcolor{red}{red} and \textcolor{blue}{blue} colors respectively represent the optimal and suboptimal results.}
 \label{tab:label_type}
 \centering
 \renewcommand\arraystretch{1.3}
 \setlength{\tabcolsep}{5.4mm}{
 \resizebox{1.0\linewidth}{!}{
\begin{tabular}{c|c|c|cccc|c}
\hline
\textbf{Method} & \textbf{Image encoder} & \textbf{Label Type} & \textbf{F-0} & \textbf{F-1} & \textbf{F-2} & \textbf{F-3} & \textbf{Mean} \\ \hline
\multirow{8}{*}{VRP-SAM} & \multirow{4}{*}{VGG-16} & \emph{point.} & 24.6 & 34.4 & 35.1 & 36.7 & 32.7 \\ 
 &  & \emph{scribble.} & 32.7 & 49.6 & 46.8 & 39.5 & 42.2 \\ 
 &  & \emph{box.} & 36.5 & 49.7 & 49.7 & 43.2 & 44.8 \\  
 &  & \emph{mask.} & 43.6 & \textcolor{blue}{51.7} & 50.0 & 46.5 & 48.0 \\ \cline{2-8} 
 & \multirow{4}{*}{ResNet-50} & \emph{point.} & 32.0 & 39.2 & 43.0 & 39.3 & 38.4 \\ 
 &  & \emph{scribble.} & 40.2 & 52.0 & 52.4 & 44.4 & 47.3 \\  
 &  & \emph{box.} & \textcolor{blue}{44.5} & 49.3 & \textcolor{blue}{55.7} & \textcolor{blue}{49.1} & \textcolor{blue}{49.7} \\ 
 &  & \emph{mask.} & \textcolor{red}{48.1} & \textcolor{red}{55.8} & \textcolor{red}{60.0} & \textcolor{red}{51.6} & \textcolor{red}{53.9} \\ \hline
\end{tabular}}}
\end{table*}

\textbf{Evaluation:} For quantitative evaluation on the employed benchmark, we employed the same approach as our training strategy to obtain visual reference prompts about point, scribble, and box. For points and scribbles, we randomly selected $(1, 20)$ as prompts. In visual experiments, we showcased the performance of VRP-SAM using only one point or scribble. When inferring with $N$ visual reference prompts, we utilize the visual reference prompt encoder to generate $N$ sets of queries. Subsequently, these $N$ sets of queries are concatenated and fed into the mask decoder.

\begin{table*}[]\footnotesize
    \caption{Ablation study on COCO-20$^i$ base set. The \textcolor{red}{red} and \textcolor{blue}{blue} colors respectively represent the optimal and suboptimal results.}
 \label{tab:base_set}
 \centering
 \renewcommand\arraystretch{1.3}
 \setlength{\tabcolsep}{5.4mm}{
 \resizebox{1.0\linewidth}{!}{
\begin{tabular}{c|c|c|ccccc|c}
\hline
\multirow{2}{*}{\textbf{Methods}} & \multirow{2}{*}{\textbf{Image encoder}} & \multirow{2}{*}{\textbf{Label type}} & \multicolumn{5}{c|}{\textbf{Base set}} & \textbf{Novel set} \\ \cline{4-9} 
 &  &  & F-0 & F-1 & F-2 & \multicolumn{1}{c|}{F-3} & Means & Means \\ \hline
\multirow{5}{*}{VRP-SAM} & \multirow{4}{*}{ResNet-50} & \emph{point.} & 30.1 & 36.8 & 44.5 & \multicolumn{1}{c|}{42.4} & 38.5 & 38.2 \\
 &  & \emph{scribble.} & 44.7 & 50.7 & 49.8 & \multicolumn{1}{c|}{53.1} & 49.6 & 47.2 \\
 &  & \emph{box.} & 46.4 & 50.4 & 49.8 & \multicolumn{1}{c|}{57.3} & 51.0 & 49.7 \\
 &  & \emph{mask.} & \textcolor{blue}{52.4} & \textcolor{blue}{56.4} & \textcolor{blue}{58.8} & \multicolumn{1}{c|}{\textcolor{blue}{61.3}} & \textcolor{blue}{57.2} & \textcolor{blue}{53.9} \\ \cline{2-9} 
 & DINOv2-B & \emph{mask.} & \textcolor{red}{66.1} & \textcolor{red}{69.0} & \textcolor{red}{70.0} & \multicolumn{1}{c|}{\textcolor{red}{65.2}} & \textcolor{red}{67.6} & \textcolor{red}{60.4} \\ \hline
\end{tabular}}}
\end{table*}

\section{Datasets Setting}
To quantify the generalization of VRP-SAM to unseen objects, we adopt the data setup of few-shot segmentation, organizing all classes from the COCO and PASCAL datasets into four folds. For each fold, PASCAL-5$^i$~\cite{shaban2017one} comprises 15 base classes for training and 5 novel classes for testing, while COCO-20$^i$~\cite{nguyen2019feature} includes 60 training base classes and 20 testing novel classes. Table~\ref{tab:data_PASCAL} and Table~\ref{tab:data_coco} provide a detailed breakdown of the testing classes for PASCAL-5$^i$ and COCO-20$^i$ in each fold, where the training classes are composed of combinations of testing classes from other folds. Additionally, to assess the performance of VRP-SAM on domain shift, we trained on COCO-20$^i$ and tested on PASCAL-5$^i$. To ensure that there is no overlap between training and testing classes, we performed a new partition of the PASCAL dataset, and Table~\ref{tab:data_coco2PASCAL} provides a detailed description of the newly segmented folds for PASCAL.

\section{More experiments}
\subsection{Comparison with the State-of-the-art}
To demonstrate the superior performance of VRP-SAM, we conducted experiments on the COCO-20$^i$ dataset, comparing it with state-of-the-art methods. As shown in Table~\ref{tab:app_sota}, we observed outstanding results when employing DINOv2-B~\cite{dinov2} as the image encoder in VRP-SAM, particularly surpassing SegGPT~\cite{wang2023seggpt} for the first time — a method utilizing an image encoder trained on in-domain datasets. Furthermore, utilizing ResNet-50 as the image encoder still achieved a noteworthy Mean IoU of 53.9, outperforming the majority of current approaches. These findings substantiate the versatility and robustness of VRP-SAM across different encoders, solidifying its superiority in visual reference segmentation tasks.

\subsection{Different label type}
We conducted ablation study on label types of reference image, exploring their impact on VRP-SAM results using both VGG-16 and ResNet-50. Table~\ref{tab:label_type} presents the findings, demonstrating a gradual performance improvement with increasing annotation precision, particularly from points to masks. Specifically, transitioning from point to mask annotations significantly boosted VRP-SAM performance by approximately 15 miou. This underscores the crucial influence of annotation granularity on VRP-SAM performance.

\subsection{Comparison on base set}
To assess the performance of VRP-SAM on trained categories, we randomly selected 1000 reference-target pairs from the training set of COCO-20$^i$. This subset was used to test VRP-SAM's performance on categories it had been trained on. We refer to the test set composed of categories used for training in each fold as the base set. The experimental results, as shown in Table~\ref{tab:base_set}, demonstrate that VRP-SAM performs better on the base set than on the novel set. This improvement is attributed to VRP-SAM acquiring specific knowledge about the base classes during training, contributing to enhanced performance on those classes. Moreover, employing DINOv2 as the image encoder significantly enhances VRP-SAM's performance on the base set. This improvement is attributed to DINOv2's feature representation being more adept at capturing intricate semantic relationships, thereby boosting the model's performance on known classes.

\begin{table}[h]
\caption{Compare with text-guided SAM on PASCAL-5$^i$ novel set. The \textcolor{red}{red} colors respectively represent the optimal results.}
  \centering
    \setlength{\tabcolsep}{5pt}
    \renewcommand{\arraystretch}{1.3}
    \vspace{-2mm}
    \setlength{\tabcolsep}{2.4mm}{
    \resizebox{1.0\linewidth}{!}{
    \footnotesize
    \begin{tabular}{l|cccc|c}
    \hline
    \textbf{model}  & \textbf{F-0} & \textbf{F-1} & \textbf{F-2} & \textbf{F-3} & \textbf{means} \\
    \hline
    \textcolor{gray}{\textit{text-guided SAM:}} &&&&\\
    FastSAM$_\textrm{{(ViT-B + YOLOv8x)}}$ & 18.9 & 29.1 & 24.4 & 32.0 & 26.1 \\
    CLIP-SAM$_\textrm{{(ViT-B + SAM-H)}}$ & 25.9 & 51.8 & 33.5 & 51.3 & 40.6 \\
    \hline
    \textcolor{gray}{\textit{visual-guided SAM:}} &&&&\\
    VRP-SAM$_\textrm{{(RN50 + SAM-H)}}$ & \textcolor{red}{73.9} & \textcolor{red}{78.3} & \textcolor{red}{70.6} & \textcolor{red}{65.1} & \textcolor{red}{71.9} \\ \hline
  \end{tabular}}
  \vspace{-1.em}
  \label{tab:text_guide}}
\end{table}

\subsection{Compare with text-guided SAM}
The text-guided SAM can also segment target objects based on category names, thus achieving a functionality similar to VRP-SAM. To verify this, we conducted a comparison with existing text-guided SAM in the PASCAL-5$^i$ setting. The experimental results, as shown in Table~\ref{tab:text_guide}. indicate that the performance of text-guided SAM is poor. This not only demonstrates that SAM is a semantically agnostic segmentation model, but also underscores the superiority of VRP-SAM.

\subsection{Results on more application scenarios}
In the aforementioned experiments focusing on typical object segmentation tasks, we conducted a comprehensive evaluation of VRP-SAM. Next, we validate the effectiveness of VRP-SAM in atypical object segmentation tasks. Specifically, we conducted experiments on both part segmentation and video object segmentation tasks. The results are shown in Table~\ref{tab:part} and ~\ref{tab:vos}. The results affirm VRP-SAM's promising performance in the domain of Part Segmentation and Video Object Segmentation. We believe these findings further support the versatility of VRP-SAM across diverse applications.

\begin{table}[h]
\caption{Result of part segmentation on PASCAL-PART. The \textcolor{red}{red} colors respectively represent the optimal results.}
  \centering
    \setlength{\tabcolsep}{2pt}
    \renewcommand{\arraystretch}{1.4}
    \footnotesize
  \vspace{-3mm}
  \setlength{\tabcolsep}{2.mm}{
    \resizebox{1.0\linewidth}{!}{
    \begin{tabular}{l|cccc|c}
    \hline
    \textbf{model}  & \textbf{animals} & \textbf{indoor} & \textbf{person} & \textbf{vehicles} & \textbf{means} \\
    \hline
    Painter~\cite{painter} & 20.2 & 49.5 & 17.6 & 34.4 & 30.4 \\
    SegGPT~\cite{wang2023seggpt} & 22.8 & 50.9 & \textcolor{red}{31.3} & \textcolor{red}{38.0} & 35.8 \\
    PerSAM~\cite{persam} & 19.9 & 51.8 & 18.6 & 32.0 & 30.1 \\
    \hline
    VRP-SAM$_{\textrm{(RN50)}}$ & 23.4 & \textcolor{red}{56.6} & 25.8 & 35.6 & 35.4  \\
    VRP-SAM$_{\textrm{(DINOv2-B)}}$ & \textcolor{red}{30.3} & 52.1 & 25.8 & 36.7 & \textcolor{red}{36.2} \\ \hline
  \end{tabular}}}
  \label{tab:part}
\end{table}

\begin{table}[h]
\caption{Result of video object segmentation on DAVIS 2017 dataset. The \textcolor{red}{red} colors respectively represent the optimal results.}
  \centering
    \setlength{\tabcolsep}{5pt}
    \renewcommand{\arraystretch}{1.3}
    \tiny
  \vspace{-3mm}
  \setlength{\tabcolsep}{4.4mm}{
    \resizebox{1.0\linewidth}{!}{
    \begin{tabular}{l|ccc}
    \hline
    \textbf{model} & \emph{J\&F} & \emph{J} & \emph{F} \\
    \hline
    Painter~\cite{painter} & 34.6 & 28.5 & 40.8 \\
    PerSAM~\cite{persam} & 60.3 & 56.6 & 63.9 \\
    \hline
     VRP-SAM & \textcolor{red}{64.8} & \textcolor{red}{62.1} & \textcolor{red}{67.4}   \\ \hline
  \end{tabular}}}
  \label{tab:vos}
\end{table}

\section{Limitation and future works}
Currently, we only demonstrate the effectiveness of VRP-SAM in few-shot semantic segmentation. Extending VRP-SAM to more vision tasks, such as video object segmentation and object tracking, needs more investigation. We leave it for future work. 

{
    \small
    \bibliographystyle{ieeenat_fullname}
    \bibliography{egbid}
}

\end{document}